\documentclass{bmvc2k}

\usepackage{times}
\usepackage{epsfig}
\usepackage{graphicx}
\usepackage{amsmath}
\usepackage{amssymb}
\usepackage{array}
\usepackage{booktabs}
\usepackage{multirow}
\usepackage{algorithm}
\usepackage{algpseudocode}
\usepackage[normalem]{ulem}
\usepackage{subfigure}
\usepackage{xspace}
\usepackage[export]{adjustbox}
\usepackage{tabularx}
\usepackage{multirow}
\usepackage{wrapfig}
\newcommand{\etal}{\textit{et al}.}

\newcommand{\eg}{\textit{e}.\textit{g}.}

\usepackage{pifont}

\newcommand{\cmark}{\ding{51}}

\definecolor{red}{rgb}{0,0,0}
\newcommand{\red}[1]{\textcolor{red}{{#1}}}

\title{Wide and Narrow: \\Video Prediction from Context and Motion}

\addauthor{Jaehoon Cho}{rehoon@yonsei.ac.kr}{1}
\addauthor{Jiyoung Lee}{easy00@yonsei.ac.kr}{1}
\addauthor{Changjae Oh}{c.oh@qmul.ac.uk}{2}
\addauthor{Wonil Song}{swonil92@yonsei.ac.kr}{1}
\addauthor{Kwanghoon Sohn}{khsohn@yonsei.ac.kr}{1}

\addinstitution{
 School of Electrical and \\
 Electronic Engineering\\
 Yonsei University\\
 Seoul, South Korea
}
\addinstitution{
 Centre for Intelligent Sensing\\
 Queen Mary University of London\\
 London, United Kingdom
}

\runninghead{J. CHO ET AL.}{Video Prediction from Context and Motion}

\def\eg{\emph{e.g}\bmvaOneDot}

\def\etal{\emph{et al}\bmvaOneDot}

\begin{document}

\maketitle

\begin{abstract}
Video prediction, forecasting the future frames from a sequence of input frames, is a challenging task since the view changes are influenced by various factors, such as the global context surrounding the scene and local motion dynamics. In this paper, we propose a new framework to integrate these complementary attributes to predict complex pixel dynamics through deep networks.
To capture the local motion pattern of objects, we devise local filter memory networks that generate adaptive filter kernels by storing the prototypical motion of moving objects in the memory. 
We further present global context propagation networks that iteratively aggregate the non-local neighboring representations to preserve the contextual information over the past frames.
The proposed framework, utilizing the outputs from both networks, can address blurry predictions and color distortion. We conduct experiments on Caltech pedestrian and UCF101 datasets, and demonstrate state-of-the-art results. Especially for multi-step prediction, we obtain an outstanding performance in quantitative and qualitative evaluation.
\end{abstract}

\section{Introduction}
Video prediction, the task of generating high-fidelity future frames by observing a sequence of past frames, is paramount in numerous computer vision applications such as video object segmentation~\cite{xu2019spatiotemporal}, robotics~\cite{finn2016unsupervised}, anomaly detection~\cite{liu2018future}, and autonomous driving applications~\cite{byeon2018contextvp,gui2018few,cho2018multi}.

	Early works addressed the video prediction to directly predict raw pixel intensities from the observations through various deep learning architectures including 2D/3D convolutional neural networks (CNNs)~\cite{mathieu2015deep,vondrick2016generating}, recurrent neural networks (RNNs)~\cite{ranzato2014video,wang2017predrnn,srivastava2015unsupervised,xingjian2015convolutional}, and generative adversarial networks (GANs)~\cite{mathieu2015deep,vondrick2016generating,kwon2019predicting,liang2017dual}.
	These methods have improved the perceptual quality by encoding the spatial contents, but they often overlook the complex motion variations over time.
	Another approach is to explicitly model the motion dynamics by predicting a dense motion field (\eg optical flow)~\cite{patraucean2015spatio,liu2017video,liang2017dual,reda2018sdc,liu2018future,li2018flow}. The dense motion field ensures the temporal consistencies between frames, but occlusion and large motion may limit precise motion estimation ~\cite{gao2019disentangling}.

	 Humans perceive an entire image, recognize the regional information such as moving objects~\cite{rensink2000dynamic}, and can predict the next scene. Likewise, recent approaches in video prediction have attempted to decompose high-dimensional videos into various factors, such as pose/content~\cite{hsieh2018learning}, motion/content~\cite{villegas2017decomposing,tulyakov2018mocogan}, motion context~\cite{Lee2020long}, multi-frequency~\cite{jin2020exploring}, and object parts~\cite{xu2019unsupervised}. The decomposed factors are easier to predict since they address the prediction on lower-dimensional representations.

 \begin{figure}[t]
 	\centering
 	\subfigure[]{\includegraphics[width=0.24\linewidth]{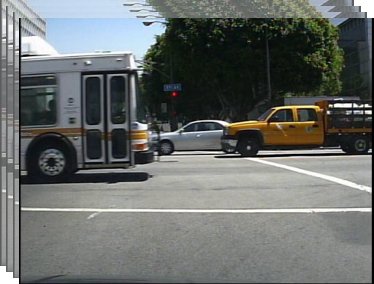}\label{fig:1a}}
 	\subfigure[]{\includegraphics[width=0.24\linewidth]{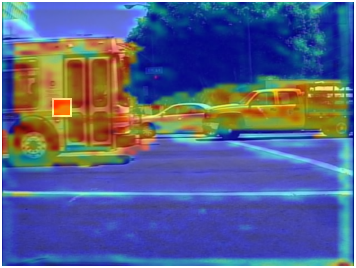}\label{fig:1b}}
 	\subfigure[]{\includegraphics[width=0.24\linewidth]{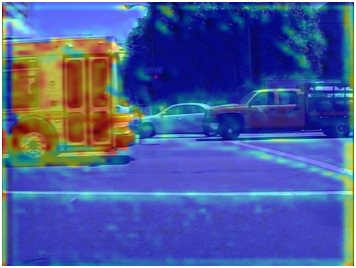}\label{fig:1c}}
	\subfigure[]{\includegraphics[width=0.24\linewidth]{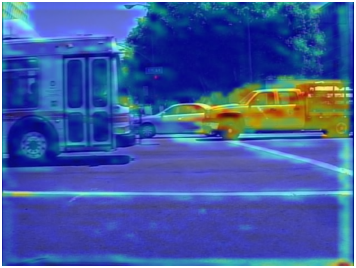}\label{fig:1d}}
 	\\ 
 	\caption{Examples of intermediate features in our network. Given a video sequence in (a), our model captures two complementary attributes, contextual information and motion dynamics. (b) Attention map for a query location (red point) from the global context propagation networks (GCPN), and (c)--(d) memory attention maps for different memory slots of the local filter memory networks (LFMN) respectively. The attention maps are upsampled to match the image size. Best viewed in color.
 	}\vspace{-15pt}
 	\label{fig:1}
 \end{figure}

  In this paper, we propose a video prediction framework that captures the global context and local motion patterns in the scene with two streams: local filter memory networks (LFMN) and global context propagation networks (GCPN). 
  We present memory-based local filters that address the motion dynamics of objects over the frames. By leveraging the memory module~\cite{weston2014memory}, LFMN can generate filter kernels that contain the prototypical motion of moving objects on the scene.
  {Namely, the filter kernels with memory facilitate learning the pixel motions (Fig.~\ref{fig:1c} and \ref{fig:1d}) and predicting long-term future frames.}  
  Second, to capture the global context that lies over the consecutive frames, GCPN iteratively propagates the context from reference pixels to the entire image by computing the pairwise similarity in a non-local manner~\cite{wang2018non}. Through the propagation steps, GCPN aggregates the regions which have similar appearances to predict future frames consistent to the global context (Fig.~\ref{fig:1b}).
  Finally, we integrate the global and local information by filtering the global feature from GCPN with memory-based kernels containing motion patterns from LFMN (Fig.~\ref{fig:output_LFMN}).

  The proposed method can preserve the global context and refine the movements of dynamic objects simultaneously, resulting in more accurate video prediction.
  In experiment, we validate our method on the Caltech pedestrian~\cite{dollar2009pedestrian} and UCF101~\cite{soomro2012ucf101} datasets. Our method shows state-of-the-art performance and improves the prediction in the large-motion sequences and multi-step prediction.

      \vspace{-13pt}

\section{Related Work}
  
  \paragraph{Video prediction.}
  
    Early studies in video prediction adopt recurrent neural networks (RNNs) to consider the temporal information between frames~\cite{ranzato2014video}.
    To achieve better performance in long-term prediction, Long-Short-Term-Memory (LSTM) networks~\cite{srivastava2015unsupervised} and convLSTM~\cite{xingjian2015convolutional} is proposed.
    Recent study~\cite{wang2019memory} exploits two cascaded adjacent recurrent states for spatio-temporal dynamics. Motivated by the recent success of 3D convolutions, a lightweight model is proposed to use a two-way autoencoder incorporating 3D convolutions~\cite{yu2019efficient}, and a model is designed to integrate 3D convolutions into RNNs~\cite{wang2018eidetic}.Several recent methods have adopted GAN since it generates sharper results in image generation~\cite{kwon2019predicting,liang2017dual,mathieu2015deep,vondrick2016generating,tulyakov2018mocogan}.
	Since they focus on generating future frames at a global level through {direct pixel synthesis}, they show limited ability in capturing moving objects and large motions.

    Another line of research computes the pixel-wise motion information from consecutive frames, and then, explicitly learning it or exploiting it as input~\cite{patraucean2015spatio,liu2017video,liang2017dual,reda2018sdc,liu2018future,li2018flow,gao2019disentangling,xu2019unsupervised}. 
    {The estimated flow information is partially utilized to predict the future movement of an object~\cite{xu2019unsupervised}.}
    Moreover, motion information can be extended to spatio-temporal domain~\cite{liu2017video} or multi-step flow prediction~\cite{li2018flow} for video prediction.
    However, these methods estimate the next frames with the local receptive fields while not fully considering the global context.
    Moreover, some studies~\cite{pottorff2019video,xu2019unsupervised,tulyakov2018mocogan,Lee2020long,park2021vid} using motion achieved limited success on few simple datasets which have low variances such as the Moving MNIST~\cite{srivastava2015unsupervised}, KTH action~\cite{schuldt2004recognizing}, and Human 3.6M~\cite{ionescu2013human3}.
    Unlike previous works, we conduct experiments on more challenging datasets including Caltech and UCF101 and exploit two complementary attributes by incorporating global contextual information and local motion dynamics. 
    
    Future prediction errors in incomplete models can be classified as follows~\cite{wang2018eidetic,oprea2020review}: (i) how to estimate the systematic errors owing to a lack of modeling ability for deterministic variations, (ii) how to model the probabilistic and inherent uncertainty about the future~\cite{xu2018video,franceschi2020stochastic,wang2020probabilistic,guen2020disentangling}. Our paper belongs to video prediction, which focuses on the first factor.

    \vspace{-8pt}
  
  \paragraph{Memory network.}
  	To address the long-term dependencies in video prediction, various methods have employed RNNs or LSTM as their backbone  models~\cite{atkeson1995memory,graves2013generating,hochreiter1997long}. However, their capacity is not large enough to accurately recover the features from the past information. To overcome the problems, the memory networks~\cite{weston2014memory} introduce an external memory component that can be read and written for prediction. 
  	The external memory can be used to address various vision problems, including video object segmentation~\cite{oh2019video,seong2020kernelized}, image generation~\cite{zhu2019dm}, object tracking~\cite{yang2018learning}, and anomaly detection~\cite{gong2019memorizing,park2020learning}.
  	In our work, we extend the idea of the memory networks to make it suitable for our task, video prediction.
  	Specifically, the memory in local networks is dynamically recorded and updated with the new prototypical motion of moving objects, which helps in predicting motion dynamics in videos.
  \vspace{-11pt}
  
   \section{Proposed Method}
  
  \subsection{Problem Statement and Overview}
  
  Let ${X}_{t}$ be the $t^{th}$ frame in the video sequence $\mathbf{X} = (X_{t-(\delta-1)},...,X_{t})$ of the last $\delta$-frames.
  {The goal of video prediction is to generate the next $n$ frames $\mathbf{Y} = (Y_{t+1}, ..., Y_{t+n})$ from the input sequence $\mathbf{X}$.}
  The major challenge in video prediction is to handle the complex evolution of pixels by composing two key attributes of the scene, \textit{i.e.}  context of the content and motion dynamics.
  
\begin{figure*}[!]
   	\centering
   	{\includegraphics[width=0.85\linewidth]{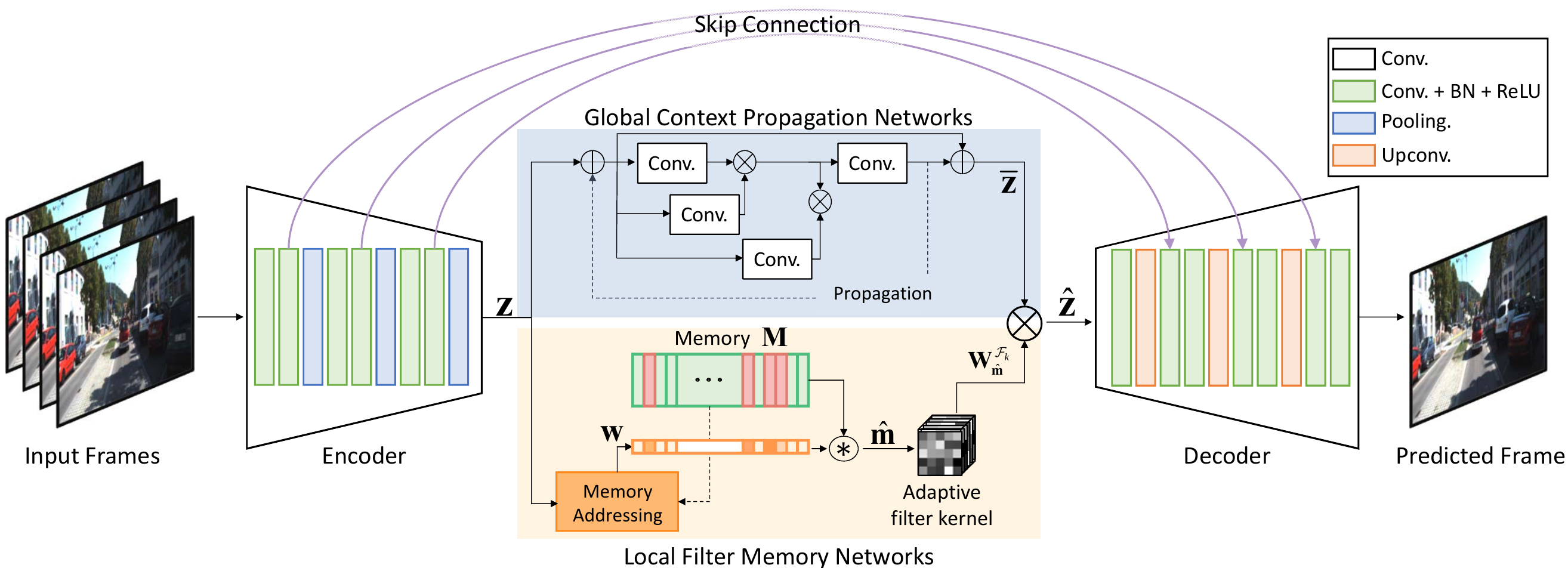}}
   	\vspace{-8pt}
   	\caption{ 
   		Overview of the proposed method. Our model consists of global context propagation networks (GCPN) and local filter memory networks (LFMN).
   		$\otimes$, $\oplus$, and $\circledast$ are the convolution, addition, and multiplication operation, respectively.
   	}
   	\label{fig:3}
   	\vspace{-11pt}
   \end{figure*}

  {To this end, as shown in Fig.~\ref{fig:3}, we devise a new framework that takes advantages of two complementary components estimated from two sub-networks: local filter memory networks (LFMN) and global context propagation networks (GCPN).}
  Following the previous works~\cite{reda2018sdc,kwon2019predicting,liu2018future}, our networks take the concatenated $\delta$ frames as inputs and feed them into the encoder to obtain high-dimensional embedded representation.
LFMN takes the encoded representation to generate adaptive filter kernels that contain the prototype of moving objects.
  GCPN transforms the encoded representation to capture the contextual information by iteratively propagating all the points based on a non-local manner~\cite{wang2018non}.
  To incorporate global context and motion dynamics effectively, the intermediate features from GCPN are convolved with the filters generated from LFMN.
  Finally, the filtered features are fed into the decoder to predict the next frame.
  \vspace{-11pt}
   
  	\subsection{Local Filter Memory Networks}

\begin{wrapfigure}{r}{0.4\textwidth} 
	\centering
	\renewcommand{\thesubfigure}{} 
	\subfigure{{\includegraphics[width=0.3\linewidth]{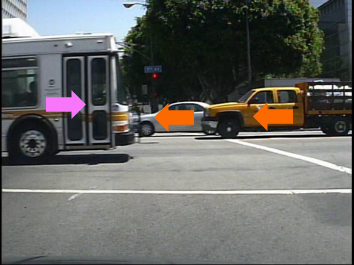}}}
	\subfigure{{\includegraphics[width=0.3\linewidth]{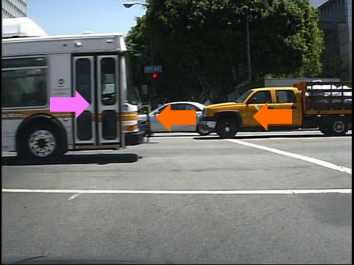}}}
	\subfigure{{\includegraphics[width=0.3\linewidth]{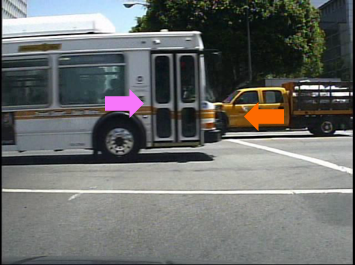}}}\\
	\vspace{-8pt}
	\subfigure{{\includegraphics[width=0.3\linewidth]{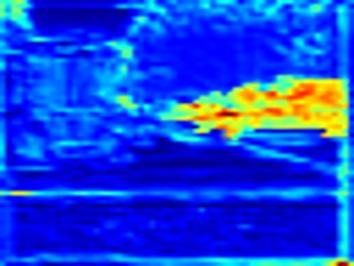}}}
	\subfigure{{\includegraphics[width=0.3\linewidth]{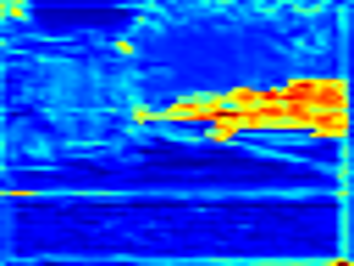}}}
	\subfigure{{\includegraphics[width=0.3\linewidth]{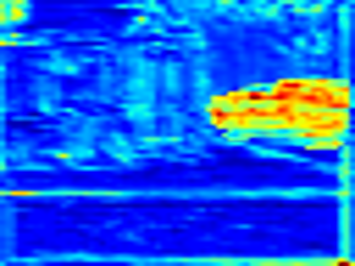}}}\\
	\vspace{-8pt}
	\subfigure{{\includegraphics[width=0.3\linewidth]{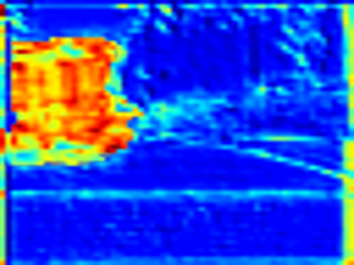}}}
	\subfigure{{\includegraphics[width=0.3\linewidth]{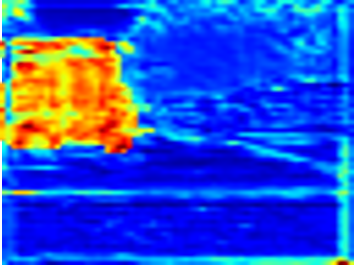}}}
	\subfigure{{\includegraphics[width=0.3\linewidth]{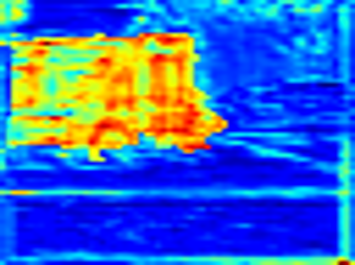}}}
	\\			
	\caption{Visualization of memory attention maps in LFMN.
		In the sequence with moving objects ($1^{st}$ row), the attention maps are activated by different memory items  with respect to the object motion ($2^{nd}$, $3^{rd}$ rows). The attention maps are upsampled to match the image size.
	}
	\label{fig:memory}\vspace{-15pt}
\end{wrapfigure}

  	{To consider motion dynamics in video sequences, we introduce LFMN that captures and memorizes the prototypical motion of moving objects from the encoded representation $\mathbf{Z}\in\mathbb{R}^{W \times H \times C}$.
	For instance, as shown in \ref{fig:memory}, the objects in the sequence can move different directions. }
  	We thus aim at learning to update and address kernel parameters encoding prototypical motion patterns that transform global representation to be robust in dynamic changes.
  	   	\vspace{-8pt}

  	We design the memory as a matrix $\mathbf{M}\in\mathbb{R}^{N \times C}$ containing $N$ real-valued memory vectors of fixed channel dimension $C$.
  	The prototypical items of the encoded future-relevant local motion features are updated to be written into the memory module.  
  	Given the encoded representation $\mathbf{Z}$, the memory networks retrieve a memory item most similar to $\mathbf{Z}$.
  	In the memory addressing operator~\cite{weston2014memory}, it takes the encoding $\mathbf{Z}$ as query to obtain the addressing weights.
  	We compute the similarity of the memory items and the encoded representation $\mathbf{Z}$.  	
  	We compute each weight $\mathbf{w}_{i}$ for $i^{th}$ memory vector $\mathbf{m}_{i}$ by applying softmax operations as follows:
  	\begin{equation}\label{eq:1}
  	\mathbf{w}_{i}= \frac{{\exp (d(\mathbf{Z},\mathbf{m}_{i}))}}{{\sum\nolimits_{j = 1}^N {\exp (d(\mathbf{Z},\mathbf{m}_{j}))} }},
  	\end{equation}
  	where $d(\cdot,\cdot)$ denotes a similarity measurement. 
  	We define $d(\cdot,\cdot)$ similar to ~\cite{santoro2016one} as follows:
  	\begin{equation}\label{eq:2}
  	d(\mathbf{Z},\mathbf{m}_{i}) = \frac{{\mathbf{Z}\mathbf{m}_{i}^{\mathsf{T}}}}{{\left\| \mathbf{Z} \right\|\left\| \mathbf{m}_{i} \right\|}}.
  	\end{equation}
  	Given an encoded representation $\mathbf{Z}$, the memory networks obtain an aggregated memory feature ${\mathbf{\hat m}} \in \mathbb{R}^{W \times H \times C}$ relying a soft addressing weight $\mathbf{w}\in\mathbb{R}^{W \times H \times N}$ as follows  
  	\begin{equation}\label{eq:3}
  	{{\mathbf{\hat m}}} = \mathbf{w}\mathbf{M} =\sum\nolimits_{i=1}^N {{\mathbf{w}_i}} {{\mathbf{m}}_{i}}.
  	\end{equation}	
  	The ${\mathbf{\hat m}}$ is a combination of the memory items most similar to $\mathbf{Z}$.    	
  	This allows our model to capture various local motion patterns using memory items.
  	Finally, we generate the future-relevant local motion filter kernels using a filter generating network $\mathcal{F}_{k}$ such that
  	\begin{equation}\label{eq:4}
  	\mathbf{W}_{\mathbf{\hat m}}^{\mathcal{F}_{k}}=\mathcal{F}_{k}({\mathbf{\hat m}}),\
  	\end{equation}
  	where $k$ is the kernel size and $\mathbf{W}_{\mathbf{\hat m}}^{\mathcal{F}_{k}} \in \mathbb{R}^{W \times H \times C \times C \times k \times k}$ denotes the generated local motion filters.
  	This generates filter kernels for each pixel on conditioned video frames.
  	While the DFN~\cite{jia2016dynamic} are limited on having to predict the next frame within the short-length input frame, our LFMN enables us to capture the long-term dynamics thanks to the memory.
  	Fig.~\ref{fig:memory} shows a visualization of an attention map that is activated by memory items.
  	The second row shows prototype patterns such as orange arrows which are activated in the movement of vehicles moving to the left. 
  	The third row shows that another type of prototype pattern is activated in the movement of objects moving to the right.
  	We can observe that that each memory item addresses the particular movement pattern of objects.
  
  	\subsection{Global Context Propagation Networks}~\label{3.4}
\vspace{-8pt}
  
  	The objective of GCPN is to capture a spatially wide range by propagating neighbor observations.
  	To capture the content from different query locations far away within the image, GCPN is built upon non-local networks~\cite{wang2018non} which calculates the pairwise relationship regardless of the pixel position.
  	However, non-local networks have a weakness to contain contextual relationships effectively~\cite{cao2019gcnet}, which may not always be effective to capture similar content information.
  	Instead of directly using non-local networks, we construct a propagation step that aggregates more relevant elements along with the most discriminative parts.
  	  
    	  	  \begin{wrapfigure}{r}{4cm}  	     	\vspace{-13pt}
  	  	  \scalebox{0.65}{
  \begin{minipage}{0.45\textwidth}
	\begin{algorithm}[H]
    \textbf{Input}: Intermediate feature $\mathbf{Z}$ \\
    \textbf{Output}: Global contextual feature $\mathbf{\bar{Z}}$ \\
    \textbf{Parameters} :{$\mathbf{W}_\theta$, $\mathbf{W}_\phi$, $\mathbf{W}_g$, $\mathbf{W}_o$}
    \begin{algorithmic}[1]
    \Procedure{Propagation}{$\mathbf{Z}$}
        \State Initialize $\mathbf{h}^0 = \mathbf{Z}$
        \For { $l = 1$ \textbf{to} $L$}
            \State $\mathbf{c} = (\mathbf{h}^{l-1}\mathbf{W}_\theta) \cdot (\mathbf{h}^{l-1}\mathbf{W}_\phi)^{\mathsf{T}}$ 
            \State $\mathbf{\bar{c}} = \text{softmax}(\mathbf{c})$ 
            \State $\mathbf{h}^{l} = \mathbf{\bar{c}} \cdot ( \mathbf{h}^{l-1}\mathbf{W}_g)$
        \EndFor
        \State $\mathbf{\bar{Z}} = \mathbf{Z} + \mathbf{h}^{L}\mathbf{W}_{o}$
    \EndProcedure
    \end{algorithmic}
    \caption{GCPN}
    \end{algorithm}
   	\vspace{-13pt}
        \end{minipage}}
  \end{wrapfigure}

Algorithm 1 summarizes GCPN.
Given the encoded feature $\mathbf{Z}$, we compute the affinity matrix representing the relationship between all points. Unlike~\cite{wang2018non}, we update the affinity matrix iteratively to propagate the future relevant context information. 
We conduct matrix multiplication between the updated affinity matrix ($h^L$) and a non-local block ($\mathbf{W}_o$) that denote linear transformation matrices (e.g., 1$\times$1 convolution).
At each step $l$, the feature is computed by aggregating the non-local neighboring representations~\cite{wang2018non}.
The global context is propagated through GCPN that computes affinity matrix via self-attention operation, $\mathcal{G}$, such that
\begin{equation}
{\mathbf{h}^{l}} = \mathcal{G}({\mathbf{h}^{l - 1}}),\\
\end{equation}
where $l={1,...,L}$ and $\mathbf{h}^{0}=\mathbf{Z}$.

At the final propagation step $L$, globally enhanced feature $\mathbf{\bar{Z}}$ is calculated by the sum of initial feature $\mathbf{Z}$ and propagated feature:
\begin{equation}
\mathbf{\bar{Z}} = \mathbf{Z} + \mathbf{h}^{L}\mathbf{W}_{o},
\end{equation}
with a weight matrix $\mathbf{W}_{o}\in\mathbb{R}^{C \times C}$.
This procedure allows GCPN to consider all positions for each location at each step and produce non-locally enhanced feature representations by propagating the future-relevant contextual information. 

	To incorporate global information and local information, we merge the output features and generated filter kernels from GCPN and LFMN by the convolutional filtering operation, respectively.
	Given the enhanced feature $\mathbf{\bar{Z}}$ from GCPN, this is convolved with the prediction of the generated filter kernels $\mathbf{W}_{\mathbf{\hat m}}^{\mathcal{F}_{k}}$ from LFMN as follows:
	\begin{equation}
	\mathbf{\hat Z} = \mathbf{\bar{Z}} \otimes \mathbf{W}_{\mathbf{\hat m}}^{\mathcal{F}_{k}},
	\end{equation}
	where $\otimes$ is the convolution operation.
	{In this process, we generate an adaptive filter that contains motion dynamic information for every pixel.}
	The resulting features are fed into the decoder and then we estimate next future frame $\hat{Y}_{t+1}$.

	        	\vspace{-11pt}

	    	\subsection{Loss Function}
	    	
	    	{To train the proposed model, we minimize an overall objective function ${\mathcal{L}}$, that includes a reconstruction loss ${\mathcal{L}}_{r}$ and a gradient loss ${\mathcal{L}_{g}}$:}
	    	
	    	\begin{equation}
	    	{{\mathcal{L}}}={{\mathcal{L}}_{r}}+\lambda_{g}{{\mathcal{L}}_{g}},
	    	\end{equation}
	    	where $\lambda_{g}$ is a weighting factor.
	    	The reconstruction loss, ${\mathcal{L}}_{r}$, measures the difference between an estimated future frame $\hat{Y}_{t+1}$ and its corresponding ground-truth frame $Y_{t+1}$:
	    	\begin{equation}
	    	{{\mathcal{L}}_{r}} = \sum\limits_{t} \left\| {{\hat Y_{t+1}} - {Y}_{t+1}} \right\|_{1},
	    	\end{equation}
	    	where {$\left\| \cdot \right\|_{1}$ is $L_{1}$ norm which does not over-penalize the error and thus enables sharper predicted images~\cite{reda2018sdc,niklaus2017video}.}
	    	We also use the gradient loss ${{\cal L}_g}$, similar to ~\cite{mathieu2015deep,liu2018future}, that computes the differences of image gradient predictions and enforces to preserve image edges effectively:
	    	\begin{multline}
	    	{{\cal L}_g} = \sum\limits_{t}\sum\limits_{u,v}\left\| {\left| {Y_{t + 1}^{u,v} - Y_{t + 1}^{u - 1,v}} \right| - \left| {\hat Y_{t + 1}^{u,v} - \hat Y_{t + 1}^{u - 1,v}} \right|} \right\|_{1}
	    	+ \left\| {\left| {Y_{t + 1}^{u,v - 1} - Y_{t + 1}^{u,v}} \right| - \left| {\hat Y_{t + 1}^{u,v - 1} - \hat Y_{t + 1}^{u,v}} \right|} \right\|_{1},
	    	\end{multline}
	    	$\hat Y_{t + 1}^{u,v}$ and $ Y_{t + 1}^{u,v}$ are the pixel elements from the estimated future frame $\hat{Y}_{t+1}$ and its corresponding ground-truth ${Y}_{t+1}$, respectively. 
	    	$u,v$ denotes the coordinates of the pixel for the width and height.
	    	$|\cdot|$ indicates the absolute value function.

    	\subsection{Implementation Details}
    	
    	All models were trained end-to-end using PyTorch~\cite{paszke2017automatic}, taking about 2 days, with an Nvidia RTX TITAN.
    	The network is trained using Adam~\cite{kingma2014adam} with an initial learning rate of 0.0002 and batch size of 16 for 60 epochs.
    	During the training, the learning rate is reduced using a cosine annealing method~\cite{loshchilov2016sgdr} and the memory $\mathbf{M}$ is randomly initialized.
    	At the testing phase, we read the memory items based on the input query.
    	All training video frames were normalised to the range of [-1, 1].
    	We set the height, $H$, width, $W$, channels, $C$, and memory items, $N$, to 64, 64, 64, and 20, respectively.
    	For the time index $t<\delta$, we copy the first frame $\delta - t$ times to get the $\delta$ frames for predicting the future frames.
    	We achieved the best results by setting the generated kernel size $k$ to 5.
    	We used a grid search to set the parameter ($\lambda_{g}$ = 0.01) on the validation set of the KITTI dataset.
    	Details of the network architecture are provided in the supplementary material.

	    \section{Experiment}
	    
	    \subsection{Experimental Settings}

	    \paragraph{Baselines.}
	    In the experiment, we consider representative baselines of the most relevant methods to our method such as PredNet~\cite{lotter2016deep}, BeyondMSE~\cite{mathieu2015deep}, Dual-GAN~\cite{liang2017dual}
	    	MCNet~\cite{villegas2017decomposing}, SDC-Net~\cite{reda2018sdc},
	    	Liu \etal~\cite{liu2018future}, CtrlGen~\cite{hao2018controllable},
	    	DPG~\cite{gao2019disentangling}, Kwon \etal~\cite{kwon2019predicting}, Jin \etal~\cite{jin2020exploring}, and CrevNet~\cite{yu2019efficient}.
	    	We used the pre-trained models provided by authors for visual comparison.
	    	We additionally obtained the results from ContextVP~\cite{byeon2018contextvp}.

	    \vspace{-8pt}
	    
	    \paragraph{Metrics.}
    	For quantitative comparison, we employ several evaluation metrics that have been used most widely for video prediction such as, Mean-Squared Error (MSE), Structural Similarity Index Measure (SSIM), and Peak Signal to Noise Ratio (PSNR).
    	Since these metrics are mostly focused on the pixel-level image quality, we additionally measure Learned Perceptual Image Patch Similarity (LPIPS)~\cite{zhang2018perceptual} {as an evaluation metric for perceptual dissimilarity.}
    	Higher values of SSIM/PSNR and lower values of LPIPS indicate better quality.

    	\paragraph{Datasets.}
    	We evaluate our method on two different datasets such as Caltech pedestrian~\cite{dollar2009pedestrian} and UCF101~\cite{soomro2012ucf101} datasets.
    	The Caltech pedestrian dataset consists of $640 \times 480$ videos taken from various driving places using vehicle-mounted cameras.
    	To validate the generalization performance, we followed experimental protocols of ~\cite{byeon2018contextvp,kwon2019predicting,lotter2016deep}, where the KITTI dataset with 41K images is used of training and the Caltech pedestrian dataset is used for testing.
    	The KITTI dataset contains $375 \times 1242 $ image sequences for driving scenes.
    	Both datasets also contain dynamic scenes since they were recorded from the moving vehicles.
    	
    	In addition, we use the UCF101 dataset that is generally used for action recognition and contains $320 \times 240$ videos focusing on human activities.
    	This dataset mainly contains the moving objects in various environments.
    	Since it contains a large amount of videos (13K videos), we use 10\% of the videos, similar to the previous works~\cite{kwon2019predicting,mathieu2015deep,byeon2018contextvp}.

	    \subsection{Ablation study}
	     \begin{wrapfigure}{r}{0.5\textwidth} \vspace{-5pt}
	     	{\includegraphics[width=0.16\textwidth,height=0.06\textheight]{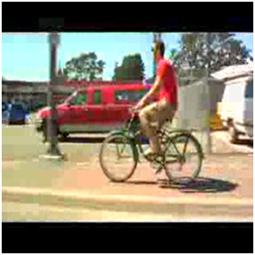}}
	     	{\includegraphics[width=0.16\textwidth,height=0.06\textheight]{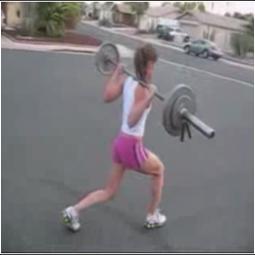}}
	     	{\includegraphics[width=0.16\textwidth,height=0.06\textheight]{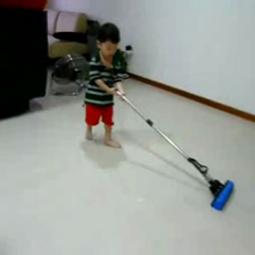}}
	     	\\
	     	{\includegraphics[width=0.16\textwidth,height=0.06\textheight]{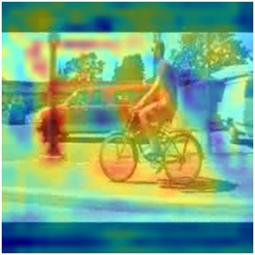}}
	     	{\includegraphics[width=0.16\textwidth,height=0.06\textheight]{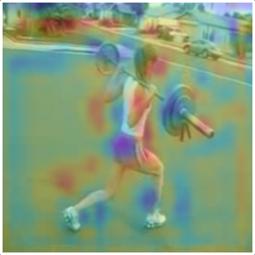}}
	     	{\includegraphics[width=0.16\textwidth,height=0.06\textheight]{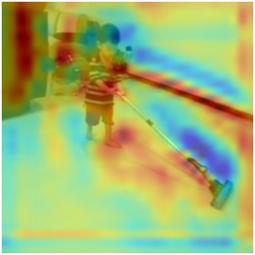}}
	     	\\
	     	{\includegraphics[width=0.16\textwidth,height=0.06\textheight]{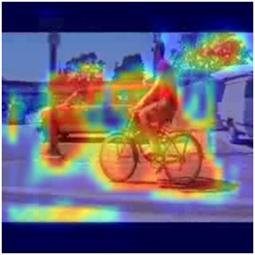}}
	     	{\includegraphics[width=0.16\textwidth,height=0.06\textheight]{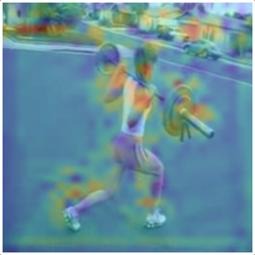}}
	     	{\includegraphics[width=0.16\textwidth,height=0.06\textheight]{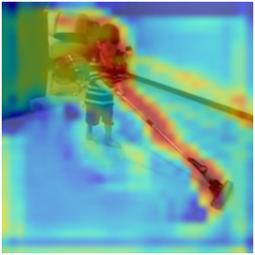}}   		
	     	\\  \vspace{-15pt}
	     	\caption{	
	     		Visualization of feature maps on the UCF101 dataset.
	     		(From top to bottom) Input frames, output features of GCPN, and output features of GCPN convolved with filter kernels from LFMN.
	     		The bottom figures show higher response around the moving objects.}
	     	\label{fig:output_LFMN}
	     \end{wrapfigure}
	    We compare the performance of our network trained with and without GCPN in Table~\ref{tab:ablation}(left).
	    The results show improved performance with GCPN as it aggregates non-local neighboring features to consider the global context effectively.
    	As shown in the third and fourth row Table~\ref{tab:ablation}(left), GCPN shows greater performance improvement than LFMN in terms of the PSNR results as GCPN improves entire images while LFMN focuses on partial details.
    	The proposed model, GCPN combined with LFMN, thus shows the best performance thanks to the benefit from both networks.

	        \begin{table}[t]
	        	\centering
	        	\scalebox{0.6}{		
	        		\begin{tabular}[b]{cc|ccc}
	        			\hline
	        			{LFMN} & {GCPN}	& {PSNR} $\uparrow$      & {SSIM} $\uparrow$ & LPIPS $\downarrow$ ($\times$ $10^{-2}$)\\
	        			\midrule
	        			\hline
	        			\ding{55} & \ding{55} & 28.2 & 0.908 &  5.08 \\ 
	        			\cmark & \ding{55} & 29.1 & 0.917 & 4.84  \\ 
	        			\ding{55} & \cmark & 29.5 & {0.919} & 4.83 \\
	        			\cmark & \cmark & \textbf{30.1} & \textbf{0.927}  & \textbf{4.81} \\
	        			\bottomrule
	        		\end{tabular}
	        	}
        	\scalebox{0.67}{		
        	\begin{tabular}[b]{c|ccccc}
				\toprule
				 {Number of memory items} & 0 & 5 & 10 & 20 & 30 \\ \cline{1-6}
				\midrule
				 PSNR $\uparrow$ & 29.5 & 29.87 & 29.92 & \textbf{30.1} & \textbf{30.1} \\
				 SSIM $\uparrow$ & 0.919 & 0.920 & 0.923 & \textbf{0.927} & 0.923 \\
				 LPIPS $\downarrow$ & 4.83 & 4.83 & 4.82 & \textbf{4.81} & \textbf{4.81} \\
				\bottomrule
			\end{tabular}
        	}	        	 \vspace{-8pt}
	        	\caption{
	        		{
	        			{Ablation studies of ours on the Caltech dataset.
	        			(Left) Results with and without LFMN and GCPN and (right) results with the different number of memory items in LFMN.}
	        		}
	        	}
	        	 \vspace{-5pt}
	        	\label{tab:ablation}
	        \end{table}



	  We compare the proposed model trained with and without LFMN as shown in Table~\ref{tab:ablation}(left).
	  Compared to the basic U-net network, a baseline, without LFMN and GCPN, our model with LFMN yields improved prediction performance in both Caltech and UCF101 datasets.
      Fig.~\ref{fig:output_LFMN} shows the examples to validate that LFMN can record the local motion patterns and generate adaptive filter kernels by combining the prototypical elements of the encoded future-relevant features.  	
	  This shows the output features of GCPN convolved with and without filter kernels from LFMN.
	  In the 3$^{rd}$ row, we show that the integration of adaptive filter kernels with the output features from GCPN.
	  With LFMN, the feature response around the moving objects becomes higher while that around the static region decreases. 
	  The results show that the output features convolved with filter kernels capture moving objects effectively.
    \begin{wrapfigure}{r}{0.5\textwidth} 	\vspace{-8pt}
    \centering
    \subfigure{\includegraphics[width=\linewidth]{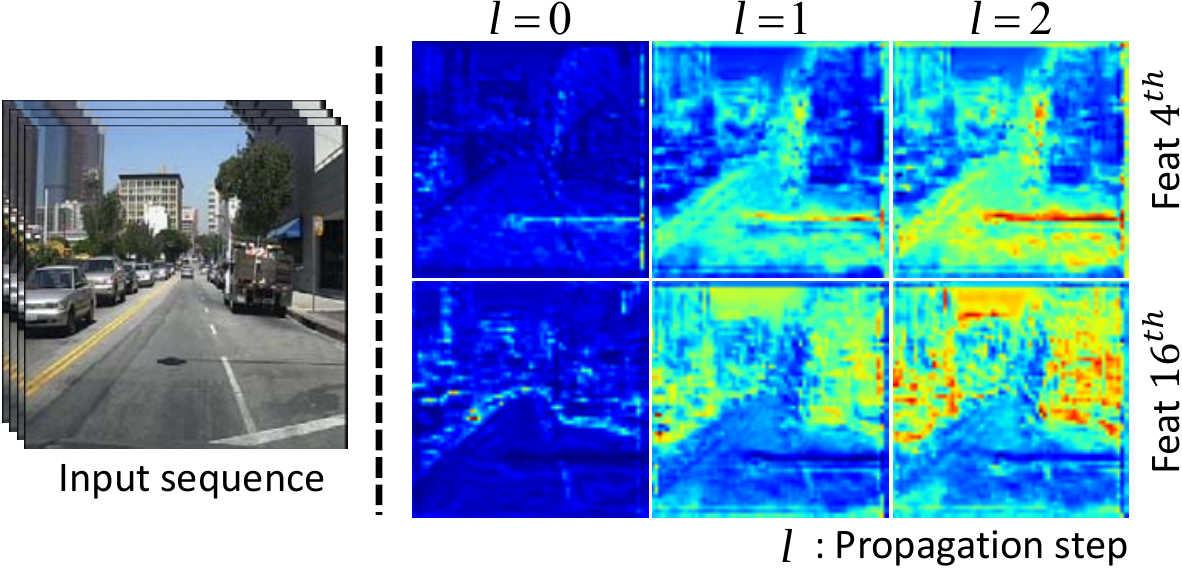}}
    \\	\vspace{-8pt}
    \caption{
    	{Visualization of feature maps of GCPN at each step $l$.
    	The feature maps are color-coded, where warmer colors indicate higher values.
    	}
    }
    \label{fig:output_GCPN}
    		\vspace{-5pt}
    \end{wrapfigure}
   
   \red{
   Table~\ref{tab:ablation}(right) shows the evaluation by varying the number of memory items ($N$), from $0$ to $30$ to verify the effect of the memory items.
    The model with a larger $N$ generally shows better prediction results since diverse prototypical motions of moving objects can be more recorded in the memory items. The effect of LFMN starts to converge when $N$ becomes $20$ or larger.}

    \red{
     To demonstrate the changes in each step $l$ in GCPN, we visualize the feature maps on the Caltech dataset, as shown in Fig.~\ref{fig:output_GCPN}.
    Fig.~\ref{fig:output_GCPN} shows that the input sequence (left) of the Caltech dataset, and the 4$^{th}$ and 16$^{th}$ feature map of GCPN (right).
	At each step $l$, GCPN aggregates non-local neighbors, propagating global spatial context.
    In addition, in terms of PSNR/SSIM, we obtained the 29.7/0.921 at $l=1$ and 30.1/0.927 at $l$=2, respectively.
    The results show the improved performance with GCPN because it aggregates non-local neighboring features to consider the global context effectively.}

		\begin{wraptable}{r}{6.5cm} {\vspace{-8pt}}
		\centering
		\scalebox{0.58}{
			\begin{tabular}[b]{c|c|ccccc}
				\toprule
				{} &  \# of frames & 2 & 4 & 6 & 8 & 10 \\ \cline{1-7}
				\midrule
				\multirow{2}*[-.3ex]{Kwon \etal~\cite{kwon2019predicting}} & PSNR $\uparrow$ & 29.17 & {29.22} & 29.01 & 28.94 & 29.01  \\
				& SSIM $\uparrow$ & 0.919 & 0.918 & {0.920} & 0.919 & 0.918  \\
				\midrule
				\multirow{2}*[-.3ex]{Ours} & PSNR $\uparrow$ & 29.73 & \textbf{30.1} & 29.51 & 29.02 & 28.93 \\
				& SSIM $\uparrow$ & 0.920 & \textbf{0.927} & \textbf{0.927} & 0.921 & 0.918 \\
				\bottomrule
			\end{tabular}
		}
		\caption{Video prediction results on the Caltech pedestrian dataset by changing the number of input frames. }
		\label{tab:2}
 		
	\end{wraptable}     
	    
   \red{
	In addition, as in ~\cite{kwon2019predicting}, we present an experiment to evaluate the effect of the number of input frames when predicting the next frame.
	Table~\ref{tab:2} shows the quantitative results according to the number of input images.
	Similar to ~\cite{kwon2019predicting}, we achieve good performance when using the inputs from 2 to 6 frames, and the performance starts to decrease when using more inputs.
    }

	     \subsection{Comparison with state-of-the-art methods}

	     \subsubsection{Next-frame prediction.}
\begin{figure*}
	        	\centering
	        	\renewcommand{\thesubfigure}{}
	        	\subfigure{\includegraphics[width=0.19\textwidth]{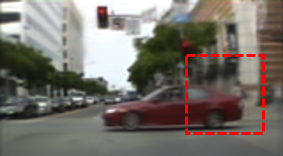}}
	        	\subfigure{\includegraphics[width=0.19\textwidth]{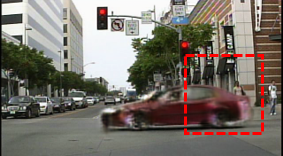}}	
	        	\subfigure{\includegraphics[width=0.19\textwidth]{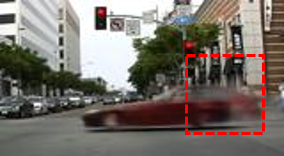}}
	        	\subfigure{\includegraphics[width=0.19\textwidth]{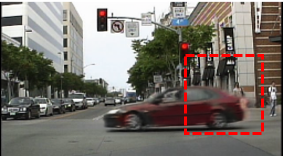}}
	        	\subfigure{\includegraphics[width=0.19\textwidth]{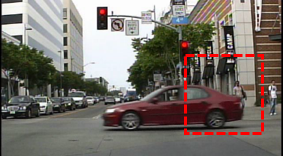}}
	        	\\ \vspace{-8pt}
	        	\subfigure[(a) PredNet~\cite{lotter2016deep}]{\includegraphics[width=0.19\textwidth,height=0.08\textheight]{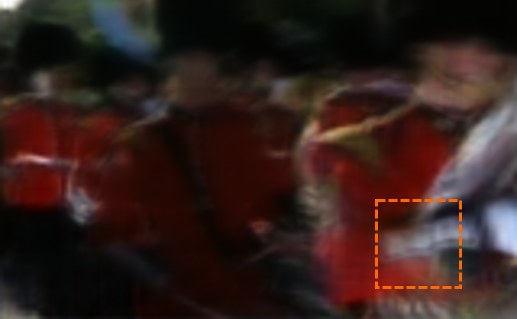}}
	        	\subfigure[(b) MCnet~\cite{villegas2017decomposing}]{\includegraphics[width=0.19\textwidth,height=0.08\textheight]{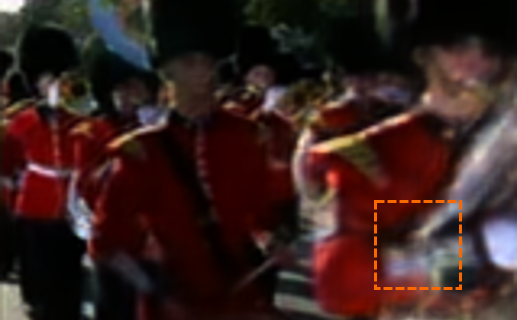}}	
	        	\subfigure[(c) ContexVP~\cite{byeon2018contextvp}]{\includegraphics[width=0.19\textwidth,height=0.08\textheight]{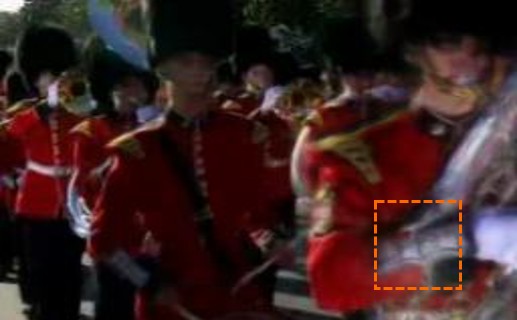}}
	        	\subfigure[(d) Ours]{\includegraphics[width=0.19\textwidth,height=0.08\textheight]{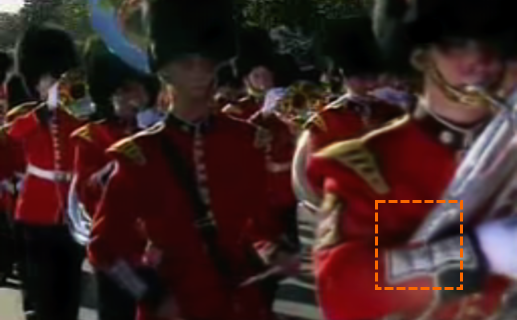}}
	        	\subfigure[(e) Ground truth]{\includegraphics[width=0.19\textwidth,height=0.08\textheight]{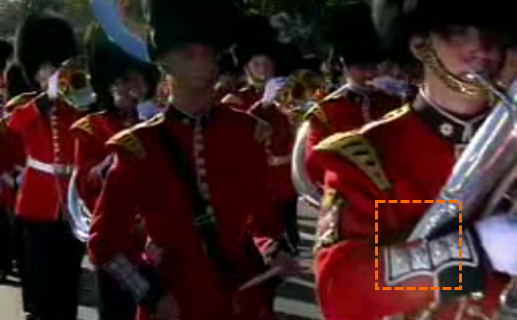}}\\ 
	        	\caption{Examples of the next-frame prediction on the Caltech Pedestrian (1$^{st}$ row) and the UCF101 dataset (2$^{nd}$ row).}
	        	\label{fig:6}
	        	\vspace{-10pt}
	        \end{figure*}

	     {To evaluate the performance of the proposed method, we compare the accuracy of the next-frame predictions with several state-of-the-art video prediction methods.
     	Sample results on thez Caltech pedestrian~\cite{dollar2009pedestrian} dataset are shown in Fig.~\ref{fig:6} in the first row.
     	Each model is only trained on the KITTI~\cite{geiger2013vision} dataset and evaluated on the Caltech pedestrian dataset without additional fine-tuning process.}
	     Our method uses 4 past frames as input images, while PredNet~\cite{lotter2016deep} and ContextVP~\cite{byeon2018contextvp} use 10 past frames as input but showing blurry results.
	     MCnet~\cite{villegas2017decomposing} also shows blurry results and unnatural deformations on the highlighted car in the first row.
	     ContextVP~\cite{byeon2018contextvp} shows a lot of artifacts in moving parts of objects such as the backside of a car.

 \begin{wraptable}{r}{6.9cm}	     \vspace{-8pt}			
	     	\centering
	     	\scalebox{0.55}{		
	     		\begin{tabular}[b]{cccccccc}
	     			\hline
	     			\multirow{3}*[-.3ex]{Method} & \multicolumn{3}{c}{\textbf{Caltech ped.}} & \multicolumn{3}{c}{\textbf{UCF101}}\\ \cmidrule(lr){2-4} \cmidrule(lr){5-7}
	     			& {PSNR $\uparrow$}      & {SSIM $\uparrow$} & MSE $\downarrow$ & {PSNR $\uparrow$} & SSIM $\uparrow$ & MSE $\downarrow$ \\		
	     			\midrule			
	     			Last Frame & 23.3 & 0.779 & 7.95 & 30.2 & 0.89 & 4.09 \\
	     			\midrule
	     			PredNet~\cite{lotter2016deep} & 27.6  & 0.91 & 2.42 & 19.87 & -  & 15.50 \\	
	     			BeyondMSE~\cite{mathieu2015deep}    & -   & 0.881 & 3.26 & 22.78  & - & 9.26 \\		
	     			Dual-GAN~\cite{liang2017dual}    & -   & 0.899 & 2.41 & 30.5  & 0.94 & -\\
	     			MCnet~\cite{villegas2017decomposing}   & -     & 0.879  & 2.50 & 31.0 &  0.91 & - \\
	     			ContexVP~\cite{byeon2018contextvp}   & 28.7   & 0.92 & 1.94 & 34.9  & 0.92 & - \\
	     			SDC-Net~\cite{reda2018sdc}    & -   & 0.918 & 1.62 & -  &  - & - \\
	     			CtrlGen~\cite{hao2018controllable}   &  26.5 & 0.90 &  - & 28.8 & 0.92  & -   \\
	     			DPG~\cite{gao2019disentangling}   &  28.2    & 0.923 &  1.62  & - & -  & -   \\
	     			Kwon \etal~\cite{kwon2019predicting}   & 29.2     & 0.919 & 1.61 & 35.0 &  0.94  & 1.37\\
	     			Jin \etal~\cite{jin2020exploring}   &  29.1    & \textbf{0.927} &  -  & - & -  & -   \\
	     			CrevNet~\cite{yu2019efficient}   &  29.3    & 0.925 & - & - & -  & - \\
	     			\midrule
         			Ours w/o ${\cal{L}}_{g}$ & 29.8 & 0.921 & 1.60 & 35.2 & 0.94 & 1.35 \\
	     			Ours   & \textbf{30.1}   &  \textbf{0.927}   & \textbf{1.58}  & \textbf{35.5}    &  \textbf{0.95}  & \textbf{1.32}  \\
	     			\bottomrule
	     		\end{tabular}
	     	}\vspace{-8pt}
	     	\caption{Quantitative evaluation of video prediction. 
	     		The table shows the performance of state-of-the-arts methods for Caltech and UCF101 datasets, respectively.
	     		MSE is averaged per pixel $( \times 10^{-3})$.
	     		The scores of other methods are adopted from original papers.
	     		The best results are presented in bold.
	     	}
	     	\label{tab:1}
	     \end{wraptable} 	   
	    
	     For capturing the motion dynamics, MCnet~\cite{villegas2017decomposing} takes the difference between two consecutive frames.
	     Because they are mainly focused on short-term information, there is a limit to their performance
	     in estimating dynamic motion.
	     In contrast, our method mitigates the above problems by considering both the global contextual information and local motion dynamics with the memory networks.
	     The second row shows the results of the UCF101~\cite{soomro2012ucf101} dataset that contains human actions taken in the wild and exhibits various challenges.
	     The results show that our method produces sharp predictions and visually pleasing results compared to the state-of-the-art methods.
	     More qualitative results are provided in supplementary materials.

	     Table~\ref{tab:1} shows a quantitative comparison with several state-of-the-art methods in both datasets.
	     We also report the results obtained by copying the last frame, which is the trivial baseline that uses the most current past frame as the prediction.
      	The last two rows of Table~\ref{tab:1} show that ${\cal{L}}_{g}$ improves the performance by preserving the image edges.
	     {
     	Our method significantly outperforms the baselines with decomposition (MCNet~\cite{villegas2017decomposing}, DPG~\cite{gao2019disentangling}, CtrlGen~\cite{hao2018controllable}) or without disentanglement.
     	For DPG~\cite{gao2019disentangling} and CtrlGen~\cite{hao2018controllable}, training with estimated optical flows may lead to erroneous supervision signals.
     	Contrarily, our method outperforms the state-of-the-art methods thanks to the GCPN, LFMN, and integration of each network.}
     	
     	\begin{wraptable}{r}{5.5cm}
		\centering
		\scalebox{0.7}{	
				\begin{tabular}[b]{c|ccc}
        			\toprule
        				&  PredNet~\cite{lotter2016deep} & MCnet~\cite{villegas2017decomposing} & Ours  \\ \hline 
        		Time & 0.321 & 0.304 & 0.451 \\
        			\bottomrule
        		\end{tabular}}\vspace{-8pt}
		\caption{Comparison of running time.}
		\label{tab:time}
		\vspace{-8pt}
	\end{wraptable}
\red{Table~\ref{tab:time} compares the running time between ours and existing methods. We follow the original setting of all the released codes. In GCPN and LFMN, we obtain 0.161s and 0.112s on average on the Caltech dataset, respectively.
On average, our method takes about 0.451.}

	     \subsubsection{{Multi-step prediction}} \label{multi}

	     To validate the time and spatial consistency in the long-term future, we present the experiment on multi-step prediction~\cite{lotter2016deep,hao2018controllable,kwon2019predicting,gao2019disentangling,villegas2017decomposing}.
	     The experimental scheme is as follows.
	      \begin{wrapfigure}{r}{0.5\textwidth}
          	\centering
          	\renewcommand{\thesubfigure}{} 
          	\subfigure[(a) PSNR]{\includegraphics[width=0.45\linewidth]{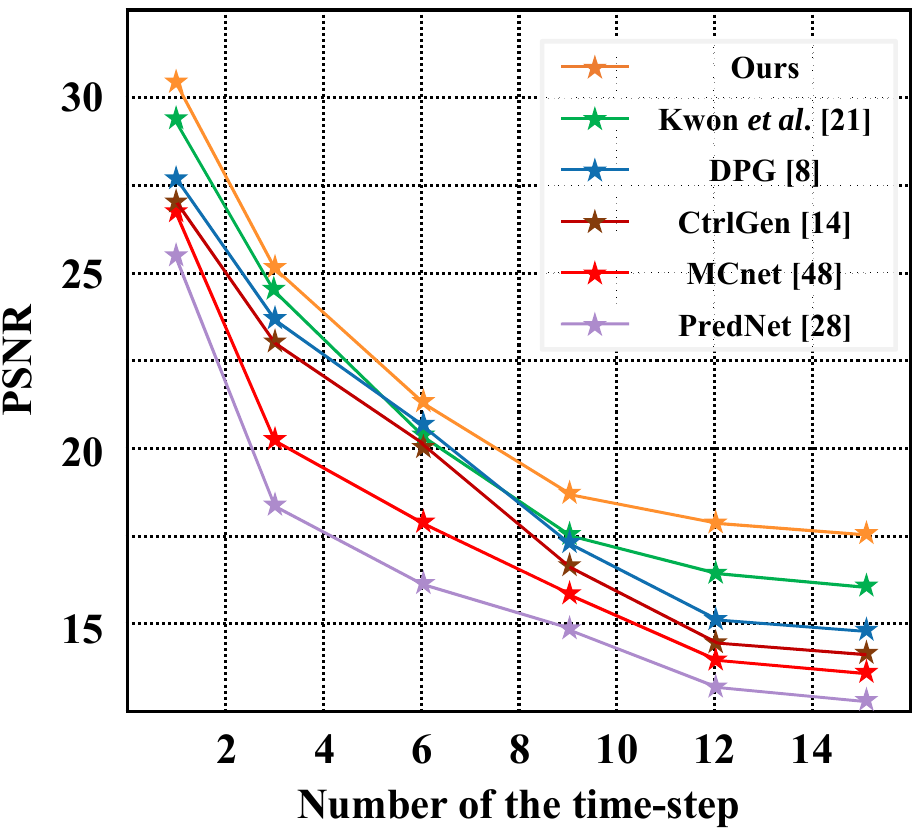}}
          	\subfigure[(b) LPIPS ($\times$ $10^{-2}$)]{\includegraphics[width=0.45\linewidth]{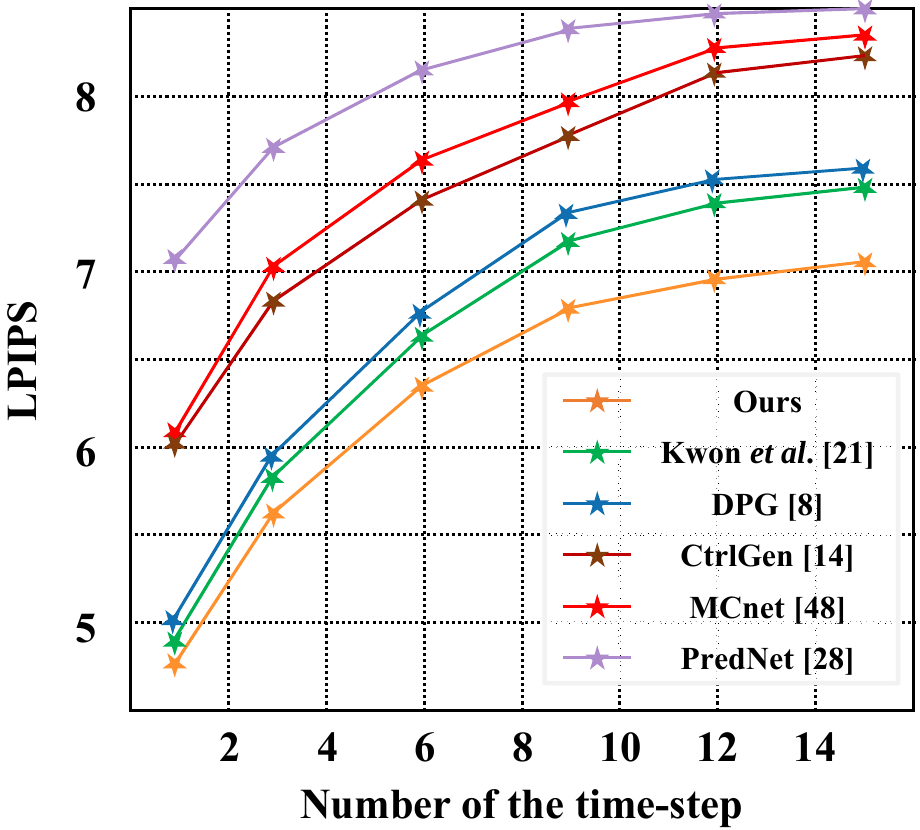}}\\
          	\caption{Quantitative comparison of multi-step prediction results with state-of-the-art methods and the proposed method on the Caltech pedestrian dataset.
          	}
          	\label{fig:7}\vspace{-8pt}
          \end{wrapfigure}
	     For example, the networks that require 4 input frames take the first 4 consecutive frames as input to find the 5$^{th}$ frame. 
	     The 6$^{th}$ frame is then predicted by using the 2$^{nd}$ to 4$^{th}$ frames and the 5$^{th}$ predicted frame as input. We continuously find the next future frame using the predicted results.
	     Fig.~\ref{fig:7} shows the quantitative results for multi-step predictions.
	     All the models take in 4 frames as input except PredNet~\cite{lotter2016deep} and recursively predict the next 15 frames.
	     Our method consistently outperforms recent approaches on all metrics over time.
	     In terms of LPIPS, while the Kwon \etal~\cite{kwon2019predicting} and DPG~\cite{gao2019disentangling} face a performance degradation significantly after 6 steps, our method does not quickly drop thanks to the previously well-predicted results.
	     The proposed method achieves outstanding results at anticipating the far future frames compared to state-of-the-art methods using with/without explicit motion estimation.
	    Due to the page limitation, we put the qualitative comparisons with state-of-the-art methods to the supplementary materials.


	    \section{Conclusion}

    	In this paper, we presented the video prediction framework that includes GCPN and LFMN, two complementary convolutional neural networks, to capture global contextual information and local motion patterns of objects, respectively.
    	GCPN considers the global context by iteratively aggregating the non-local neighboring representations.
    	LFMN generates adaptive filter kernels using memory items that have the prototypical motion of moving objects.
    	The integration of the two networks preserves the global context and refines the movements of dynamic objects simultaneously.
    	Our in-depth analysis shows that the proposed method significantly outperforms state-of-the-art methods 
    	on the next frame prediction as well as the multi-step prediction.
    	In future work, we will investigate the applicability of our model to other applications such as video semantic segmentation and anomaly detection.
  
  \clearpage
  	    \section*{Acknowledgements}
    This work was supported by the National Research Foundation of Korea (NRF) grant funded by the Korea government (MSIP) (NRF-2021R1A2C2006703) and was supported by the Yonsei University Research Fund of 2021 (2021-22-0001).
  
\bibliography{egbib}
\end{document}